\documentclass[letterpaper, 10 pt, journal, twoside]{IEEEtran}

\usepackage{booktabs,caption}
\usepackage{hyperref}
\usepackage{booktabs}
\usepackage{graphicx} % for pdf, bitmapped graphics files
\usepackage{epsfig} 
\usepackage{amsmath} % assumes amsmath package installed
\usepackage{amssymb}  % assumes amsmath package installed
\usepackage{gensymb}
\usepackage{xcolor}
\usepackage{array}
\usepackage{rotating}
\usepackage{tabularray}
\definecolor{blue}{rgb}{0,0,0}
\usepackage{pdfpages}
\usepackage{dblfloatfix}
\usepackage{multirow}
\usepackage[switch]{lineno}
\usepackage{cite}
\begin{document}

\title{Attention-Based Convolutional Neural Network Model for Human Lower Limb Activity Recognition using sEMG}

\author{M. Mollahossein$^{1}$, F.H. Daryakenari$^{2}$, M.H. Rohban$^{3}$, G.R. Vossoughi$^{*1}$
% \thanks{Manuscript received: ------, --, 2025; Revised ------, --, 2025; Accepted -------, --, 2025}
% \thanks{This paper was recommended for publication by Editor ------------ upon evaluation of the Associate Editor and Reviewers' comments.} 
\thanks{$^1$Faculty of Mechanical Engineering Department, Sharif University of Technology, Tehran, Iran. {\tt\footnotesize mojtaba.hoseini1390@yahoo.com},  {\tt\footnotesize vossough@sharif.edu}; $^2$Faculty of Engineering and IT, the University of Melbourne, Victoria, Australia. {\tt\footnotesize f.haghgoodaryakenari@unimelb.edu.au}; $^3$Computer Engineering Department, Sharif University of Technology, Tehran, Iran  {\tt\footnotesize rohban@sharif.edu}}
%$^*$equal contribution. 
\thanks{Digital Object Identifier (DOI): see top of this page.}
}

% \markboth{IEEE Robotics and Automation Letters. Preprint Version. Accepted ----------, 2025}
%{Eden \MakeLowercase{\textit{et al.}}: Comparison of Solo and Collaboration Trimanual Operation} 

\maketitle
%\thispagestyle{empty}
%\pagestyle{empty}

%%%%%%%%%%%%%%%%%%%%%%%%%%%%%%%%%%%%%%%%%%%%%%%%%%%%%%%%%%%%%%%%%%%%%%%%%%%%%%%%
\begin{abstract}
Accurate classification of lower limb movements using surface electromyography (sEMG) signals plays a crucial role in assistive robotics and rehabilitation systems. In this study, we present a lightweight attention-based deep neural network (DNN) for real-time movement classification using multi-channel sEMG data from the publicly available BASAN dataset. The proposed model consists of only 62,876 parameters and is designed without the need for computationally expensive preprocessing, making it suitable for real-time deployment. We employed a leave-one-out validation strategy to ensure generalizability across subjects, and evaluated the model on three movement classes: walking, standing with knee flexion, and sitting with knee extension. The network achieved 86.74\% accuracy on the validation set and 85.38\% on the test set, demonstrating strong classification performance under realistic conditions. Comparative analysis with existing models in the literature highlights the efficiency and effectiveness of our approach, especially in scenarios where computational cost and real-time response are critical. The results indicate that the proposed model is a promising candidate for integration into upper-level controllers in human-robot interaction systems.

\end{abstract}

\IEEEpeerreviewmaketitle
%\vspace{-5pt}
%%%%%%%%%%%%%%%%%%%%%%%%%%%%%%%%%%%%%%%%%%%%%%%%%%%%%%%%%%%%%%%%%%%%%%%%%%%%%%%%
%\linenumbers

% ///////////////////////////////////////////\\\\\\\\\\\\\\\\\\\\\\\\\\\\\\\\\\\\\\\\\\\\\\\\\
\section{Introduction}
\IEEEPARstart{S}urface Electromyography (sEMG) signals have been widely utilized in various applications, including human-machine interaction, neuromuscular disease diagnosis, and rehabilitation. One of their most significant applications is lower limb activity recognition, which plays a crucial role in healthcare monitoring, active and assistive systems, and tele-immersion \cite{Ranasinghe2016}. Human activity recognition can be performed using either visual or wearable sensors \cite{Liu2016}. Wearable sensors are placed directly on the subject’s body, whereas visual sensors, such as cameras, do not require physical attachment. However, wearable sensors are often preferred due to privacy concerns associated with visual sensors \cite{Lara2013}. With advancements in wearable sensor technology, various devices, such as accelerometers, EMG electrodes, gyroscopes, and barometers, have become available for human motion analysis. Among these, EMG sensors are particularly advantageous as they provide direct insights into muscle activation, allowing for more accurate movement recognition compared to other wearable sensors.

However, sEMG signals are inherently complex due to their intrinsic characteristics and susceptibility to various sources of noise. These signals are typically acquired using surface electrodes placed on the skin, which leads to indirect measurement of the underlying muscle activity. This indirect nature makes the signals highly vulnerable to noise, particularly from motion artifacts caused by muscle movement during activity, as well as muscle cross-talk—a result of the close proximity of adjacent muscles.

In recent years, human activity classification using sEMG signals has gained increasing attention. However, the presence of noise in EMG signals remains a significant challenge. Common noise sources include ambient interference, motion artifacts, signal instability, and cross-talk between muscle groups. While certain preprocessing techniques can help reduce noise[4], advanced noise cancellation methods are often necessary to achieve high classification accuracy. Several studies have employed various techniques to reduce noise in sEMG signals, including wavelet denoising \cite{Vijayvargiya2021}, Independent Component Analysis (ICA) , and Empirical Mode Decomposition (EMD) \cite{Vijayvargiya2022}. After denoising, the processed signals are typically fed into a classifier. However, some studies opt to use raw sEMG data (including noise) to reduce computational complexity and processing time [8]. This approach makes the classifier more suitable for real-time applications, as it eliminates the need for a time-consuming preprocessing stage. To do this, a classification modeling method that is robust to noise is needed.

With recent advancements in deep learning (DL), DL architectures have emerged as effective tools for designing robust high-level controllers based on sEMG signals \cite{Daryakenari2022}. As demonstrated in our previous work \cite{Daryakenari2022}, Convolutional Neural Networks (CNNs) \cite{Fukushima1980} are particularly well-suited for this purpose, due to their ability to automatically extract meaningful features from raw sensory data. Although CNNs are traditionally used for processing two-dimensional data such as images, they have also proven highly effective for analyzing one-dimensional signals, including sEMG.

Even in applications where the primary goal is regression, such as predicting a system's next state, CNNs can extract relevant features directly from raw input signals without the need for handcrafted preprocessing. For instance, in \cite{Gautam2020}, raw sEMG signals are first processed by a CNN, which extracts spatial features before feeding them into a Recurrent Neural Network (RNN) to predict joint angles. This highlights the versatility of CNN-based architectures for both classification and regression tasks in human-robot interaction and control systems.

Although deep learning (DL) methods have shown great potential, real-time applications require a trade-off between model accuracy and computational efficiency. In many cases, lower-accuracy models are preferred over highly accurate but computationally expensive ones, particularly for real-time scenarios. To enhance the classification performance of DL methods, dimensionality reduction techniques are often applied before feeding sEMG signals into neural networks. These techniques help extract lower-dimensional informative features, improving both accuracy and efficiency. Some commonly used methods include Linear Discriminant Analysis (LDA), Principal Component Analysis (PCA), Locally Linear Embedding (LLE), Rank-Preserving Discriminant Analysis, and Laplacian Eigenmaps \cite{Vijayvargiya2022}.

One of the key advantages of DL techniques is their ability to automate both feature extraction and dimensionality reduction, eliminating the need for manual preprocessing. While one might argue that deep learning models are more complex than shallow networks, they offer improved robustness by learning high-level representations directly from raw signals. Notably, CNNs \cite{Vijayvargiya2021}, Long Short-Term Memory (LSTM) networks \cite{Hochreiter1997}, and Deep Belief Networks (DBNs) \cite{Mohamed2009} have all been utilized for action recognition using sEMG signals [4], demonstrating promising results in this field. In addition to these well-known DL architectures, Attention Mechanisms were able to create a intra and inter data relation in DL methods. The Attention Mechanism (AM) \cite{Vaswani2017} is an additional layer in neural networks that enhances model performance by selectively focusing on the most relevant features. When strategically placed within the network, AM can improve training efficiency and, in some cases, enhance interpretability \cite{BASAN}.

With all the advances in incorporating DL modeling methods into human activity recognition, there still exists a major vague spot in previous studies. This vague spot is that in most of the previous studies they often fail to specify whether validation and test datasets contain subjects seen during the training phase. When a model is evaluated on completely unseen subjects, performance can degrade significantly. One approach to address this issue is the leave-one-out method \cite{Foroutannia2022}, where one subject is excluded from the training set and used exclusively for testing. This method provides a more realistic assessment of a model’s generalizability to new subjects.

In this study, we aim to design an Attention-Based Deep Neural Network with a low number of parameters, making it efficient for real-time applications while achieving acceptable accuracy on validation and test sets. To accomplish this, we will use an open-source dataset containing nomal subject and subjects with lower-limb abnormalities to train our model for three classes of human movement. For training, validation, and testing, we will employ a leave-one-out strategy to ensure the reported results are as generalizable as possible. Finally, our method's performance will be compared with similar approaches.
% \\\\\\\\\\\\\\\\\\\\\\\\\\\\\\\\\\\\\\\\\\\/////////////////////////////////////////////////
% ///////////////////////////////////////////\\\\\\\\\\\\\\\\\\\\\\\\\\\\\\\\\\\\\\\\\\\\\\\\\
\section{method}
\subsection{Overview of the method}

In this study, our goal is to propose a lightweight attention-based deep neural network for classifying lower limb movements using surface electromyography (sEMG) signals. To do this, we begin by utilizing the open-source BASAN dataset, which contains multi-channel sEMG and knee joint angle data collected from both healthy individuals and subjects with knee abnormalities. Minimal preprocessing is applied to the raw signals, limited to reshaping the data for neural network input and segmenting it using a sliding window approach. The data is partitioned using a leave-one-out strategy to ensure robust and generalizable evaluation. The proposed network architecture consists of multiple convolutional layers with max-pooling, followed by flattening and attention mechanisms, leading to dense layers for final classification. No additional signal denoising or feature extraction is applied, ensuring compatibility with real-time applications. At the end, our proposed model is evaluated using various metrics to assess its accuracy, efficiency, and suitability for deployment in assistive and rehabilitation technologies.

\subsection{Dataset} \label{Dataset}

The open-source BASAN dataset \cite{BASAN} includes four-channel surface electromyography (sEMG) signals and one-channel knee angle data recorded from the lower limbs of 22 male subjects aged 18 years and older, during three lower limb activities: walking on level ground (gait), standing with knee flexion, and sitting with knee extension. This dataset consists of 11 healthy individuals and 11 subjects with knee abnormalities—including six with anterior cruciate ligament (ACL) injuries, four with meniscus injuries, and one with sciatic nerve injury. sEMG signals were recorded using the Datalog MWX8, while knee joint angle data were collected using the SG150B goniometer. The data were sampled at 1000 Hz with a 14-bit resolution. After acquisition, the sEMG signals underwent band-pass filtering between 20 Hz and 460 Hz. The four sEMG electrodes were placed on key muscles involved in knee joint flexion and extension: Vastus Medialis (VM), Semitendinosus (ST), Biceps Femoris (BF), and Rectus Femoris (RF). For the experiment, the left leg of healthy subjects and the affected limb of individuals with knee abnormalities were selected. We need to highlight that the BASAN dataset does not include transition-phase data, such as sitting-to-stand movements.

\subsection{Preprocessing}
\begin{figure}
    % \centering
    \includegraphics[width=1\linewidth]{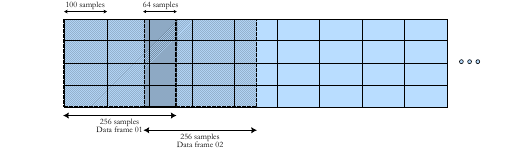}
    \caption{Data frame extraction from 4 channels of sEMG signals. The first two data frames are shown in the shaded form.}
    \label{fig:dataframe-extraction}
\end{figure}

In this work, we limit data preprocessing to the steps already performed on the dataset, as described in Section \ref{Dataset}. The only additional processing involves structuring the data appropriately for input into the neural network. To ensure proper generalization, we employ a leave-one-out method for dataset partitioning. From the 22 total participants, we allocate 9 healthy and 9 abnormal subjects for training, 1 healthy and 1 abnormal subject for validation, and 1 healthy and 1 abnormal subject for testing. Next, to format the data as neural network input, we apply a sliding window approach with a 256-millisecond window length (corresponding to 256 data samples at a 1000 Hz sampling rate) and a 64-millisecond overlap, as illustrated in Fig. \ref{fig:dataframe-extraction}. Finally, the training data is shuffled to prevent any bias toward specific subjects or groups.
%%%%%%%%%%%%%%%%%%%%%%%%%%%%%%%%%% FIGURE
%%%%%%%%%%%%%%%%%%%%%%%%%%%%%%%%%%%%%%%%%%%%%%%%%%%%%%%%%%%%%%%%%%%%%%%%%%%%
%%%%%%%%%%%%%%%%%%%%%%%%%%%%%%%%%% FIGURE
\begin{figure*}[t!]
    \centering
    \includegraphics[width=1\linewidth]{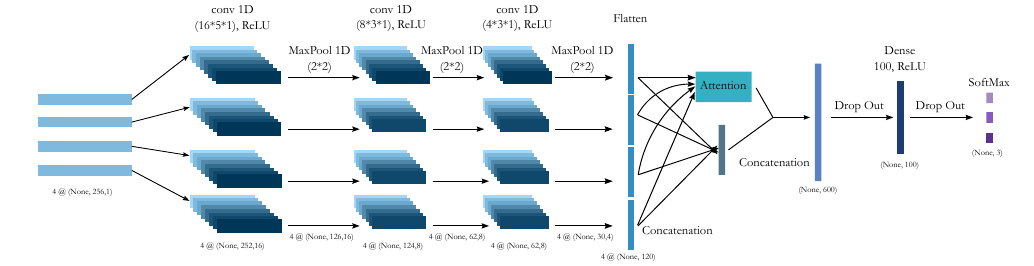}
    \caption{Network architecture.}
    \label{fig:Network_Architecture}
\end{figure*}
%%%%%%%%%%%%%%%%%%%%%%%%%%%%%%%%%%%%%%%%%%%%%%%%%%%%%%%%%%%%%%%%%%%%%%%%%%%%

\subsection{Neural Network Architecture}
The proposed neural network begins with an input layer that receives four inputs, corresponding to four sEMG channels, each with a sequence length of 256. These inputs are then processed through three consecutive convolutional layers, each followed by a max-pooling layer. The first convolutional layer employs a 5×5 kernel with 16 channels, the second uses a 3×3 kernel with 8 channels, and the third applies a 3×3 kernel with 4 channels. In all three stages, max-pooling is performed using a window size of 2. Following feature extraction, each channel's output is flattened into a one-dimensional signal. These four one-dimensional signals are then passed through two layers, which transform them into two separate one-dimensional signals. These two signals are concatenated to form a single one-dimensional representation. The resulting signal is then processed through a dense layer containing 100 neurons, followed by a final dense layer with three neurons, responsible for classification. 

The attention mechanism employed in this study is designed to enhance feature extraction from each individual sEMG channel within each input window. By focusing on the most informative aspects of the signal, the attention mechanism enables the network to emphasize relevant patterns that contribute to accurate classification. This mechanism is used in conjunction with a concatenation layer, which acts as a direct feature pathway, allowing the network to retain and combine low-level and high-level features effectively. Together, these components improve the network’s ability to deliver rich and discriminative features to the subsequent layers, thereby enhancing classification performance.

Specifically, a Bahdanau attention mechanism is implemented, where attention weights are learned through a dedicated feed-forward neural network. This allows the model to dynamically adjust its focus across the input features, learning to prioritize the most relevant temporal patterns for each classification task.

To prevent overfitting, dropout layers with a rate of 0.5 are applied before the first dense layer and after the final dense layer. The ReLU activation function is used throughout the network, except in the AM layer, which employs the tanh activation function, and in the final dense layer, where the softmax activation function is used for classification. The overall structure of the proposed network is illustrated in Figure \ref{fig:Network_Architecture}. This combination of layers led to a network that has only 62,876 parameters.
% \\\\\\\\\\\\\\\\\\\\\\\\\\\\\\\\\\\\\\\\\\\\\\//////////////////////////////////////////////
% //////////////////////////////////////////////\\\\\\\\\\\\\\\\\\\\\\\\\\\\\\\\\\\\\\\\\\\\\\

\section{Results and Discussion}

\begin{figure*}[t!]
    \centering
    \includegraphics[width=1\linewidth]{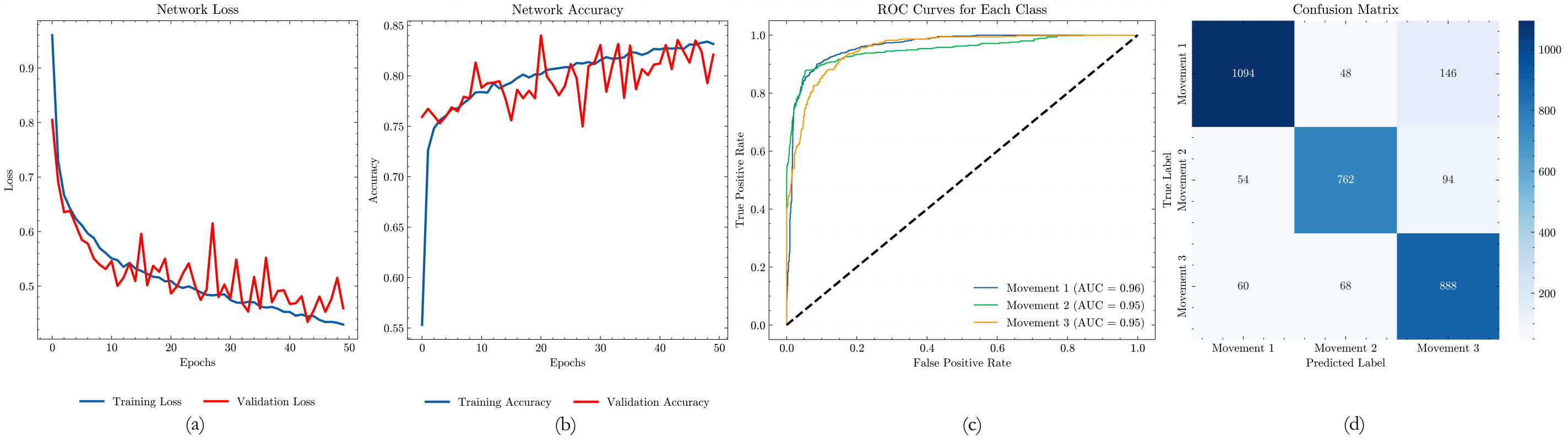}
    \caption{Results of the proposed classification network: a) Network training loss (blue line) and validation loss (red line) for 50 epochs, b) Network training accuracy (blue line) and validation accuracy (red line) for 50 epochs, c) ROC curves for each class of activity, and d) Confusion matrix}
    \label{fig:results}
\end{figure*}

The network was trained for 50 epochs, and the training and validation loss results are presented in Figure \ref{fig:results}(a). The trend observed in both training and validation curves indicates that the proposed deep neural network (DNN) successfully learned a proper mapping during the training process, with no evident signs of underfitting or overfitting. A similar trend is observed for training and validation accuracy throughout the learning process, as shown in Figure \ref{fig:results}(b). At the end of 50 epochs, the training and validation accuracy reached approximately 85\%, further confirming the model’s ability to generalize well.

To assess the performance of the model beyond the training set, the network was validated in test subjects. The Receiver Operating Characteristic (ROC) curves for each movement class are illustrated in Figure \ref{fig:results}(c). The Area Under the Curve (AUC) values were 0.96, 0.95, and 0.95, demonstrating strong classification performance. A higher AUC, closer to 1.0, indicates better classification confidence, suggesting that the proposed DNN effectively differentiates between movement patterns while minimizing false positives.

For a better understanding of misclassifications, we also analyzed the confusion matrix, as shown in Figure \ref{fig:results}(d). In this matrix, the diagonal elements represent correctly classified instances, while the off-diagonal elements indicate misclassified cases. Examining the results, movement class 1 was correctly classified in most instances; however, out of 1288 cases, it was misclassified as movement class 2 in 48 cases and as movement class 3 in 146 cases. Similarly, for movement class 2, out of 910 cases, 54 were misclassified as movement class 1, and 94 were misclassified as movement class 3. Lastly, for movement class 3, out of 1016 cases, 60 were misclassified as movement class 1, and 68 were misclassified as movement class 2. These misclassifications suggest that the movement patterns share overlapping features in the sEMG signal space, leading to occasional ambiguity in classification. This overlap highlights the inherent complexity of distinguishing certain movements solely on the basis of raw sEMG signals.

On the test subjects, the proposed model achieved an overall accuracy of 85.38\%, demonstrating strong performance in distinguishing between the three movement classes. The classification results for each movement class are summarized in Table \ref{tab:stat-results}. A class-wise evaluation (based on \ref{tab:stat-results}) reveals that Class 1 had the highest precision (90.56\%), indicating that predictions for this movement were highly reliable with minimal false positives. However, its recall (84.94\%) suggests that some instances of this class were misclassified as other movements. Class 2 exhibited a balanced performance, with precision of 86.79\% and recall of 83.74\%, highlighting the model’s ability to generalize well across this category. In contrast, Class 3 had the lowest precision (78.72\%) but the highest recall (87.40\%), implying that while the model effectively identifies most Class 3 movements, it tends to misclassify other movements as Class 3 more frequently. 

A similar result can be concluded from the F1-score, which provides a balanced measure of precision and recall. F1-score was computed for each class to evaluate the overall classification performance. As presented in Table \ref{tab:stat-results}, Class 1 achieved the highest F1-score (87.66\%), indicating that the model effectively detects and classifies this movement with minimal false positives and false negatives. Class 2 attained a slightly lower F1-score (85.23\%), suggesting that while the model performs well, there is a slight trade-off between precision and recall in this category. Class 3, however, recorded the lowest F1-score (82.84\%), primarily due to its lower precision (78.72\%) compared to its recall (87.40\%). Overall, these results highlight strong classification performance across all movement classes. 

This classification challenge likely stems from the fundamental differences between Classes 1 and 2 compared to Class 3, as Class 3 corresponds to walking, which involves more muscle activation and consequently leads to greater sEMG signal variability. These findings suggest that while the model performs well overall, future improvements could focus on reducing false positives in Class 3, potentially through additional feature extraction techniques or class-specific training strategies to enhance the model’s discriminative ability.
%%%%%%%%%%%%%%%%%%%%%%%%%%%%%%%%%% FIGURE
%%%%%%%%%%%%%%%%%%%%%%%%%%%%%%%%%%%%%%%%%%%%%%%%%%%%%%%%%%%%%%%%%%%%%%%%%%%%
%%%%%%%%%%%%%%%%%%%%%%%%%%%%%%%%%% TABLE
\begin{table}
 \caption{Prediction results of test data on each class.}
  \centering
  \begin{tabular}{llll}
    \toprule
    \multicolumn{4}{c}{Metrices}                   \\
    \cmidrule(r){2-4}
    Class          & Precision(\%)       & Recall(\%)       & F1-score(\%) \\
    \midrule
    Movement 1     & 90.56               & 84.94            & 87.66  \\
    Movement 2     & 86.79               & 83.74            & 85.23  \\
    Movement 3     & 78.72               & 87.40            & 82.83   \\
    \bottomrule
  \end{tabular}
  \label{tab:stat-results}
\end{table}
%%%%%%%%%%%%%%%%%%%%%%%%%%%%%%%%%%%%%%%%%%%%%%%%%%%%%%%%%%%%%%%%%%%%%%%%%%%%

To further evaluate the classification performance, we computed the balanced accuracy on the unseen test subjects, which was 85.36\%. Balanced accuracy accounts for class imbalances by averaging recall across all classes, ensuring that the model performs consistently across different movement types. This result closely aligns with the overall accuracy (85.38\%), indicating that the model does not exhibit bias toward any specific class.

Additionally, to support our hypothesis of designing a lightweight network, we assessed the computational complexity by measuring the inference time—the time required for the model to make a single prediction. The proposed model achieved an average inference time of less than 1 ms per sample, demonstrating its suitability for real-time applications. This efficiency makes the model highly viable for deployment in prosthetic control or rehabilitation systems, where rapid and reliable movement classification is essential.

Compared to the model presented in \cite{Gautam2020}, our network exhibits a lower accuracy; however, this difference can be attributed to several factors. One key distinction is the leave-one-out validation strategy employed in our study, which enhances the generalizability of the model. In contrast, \cite{Gautam2020} partitioned the dataset into training, validation, and test sets as a whole, meaning that the validation and test subjects, along with their related data, may have been partially seen by the network during training. This likely contributed to higher classification accuracy in their results. While treating the dataset as a whole is a common practice, the leave-one-out approach ensures a more robust evaluation by preventing data leakage and leading to more generalizable outcomes.

Additionally, our approach does not incorporate any preprocessing techniques, whereas \cite{Gautam2020} utilizes wavelet denoising. While wavelet denoising can enhance signal quality, it also introduces high computational costs, making the network less feasible for real-time applications. The additional processing required for denoising can lead to latency, which is a critical limitation in real-time prosthetic control and rehabilitation systems. In contrast, our lightweight model is designed to prioritize efficiency, ensuring minimal computational overhead and faster response times suitable for real-world applications.

 The classification method proposed in \cite{Vijayvargiya2021} shares several similarities with our approach, making it a relevant baseline for comparison. Although their method incorporates wavelet denoising as well, the data partitioning strategy is similar to ours, allowing for a direct performance comparison with some small modifications to our proposed approach. One key difference is that \cite{Vijayvargiya2021} employs a Majority-Voting-based approach for classification. In this method, four 1D CNN networks operate in parallel, each generating an independent classification result, and a voting mechanism determines the final outcome. While this technique improves reliability, we opted not to use it in our proposed DNN due to its high computational cost, particularly in real-time applications, where running four DNNs in parallel significantly increases processing time and resource consumption.

Table \ref{tab:comparison-wout-denoise} presents the classification accuracy of the proposed networks in \cite{Vijayvargiya2021}, comparing the Majority-Voting accuracy with our leave-one-out approach, without denoising. The results demonstrate that our proposed network outperforms the Majority-Voting-based network while maintaining a lower number of parameters, highlighting its computational efficiency.

To further analyze the impact of wavelet denoising, we applied the same denoising technique described in \cite{Vijayvargiya2021} to our network. The classification results, including those for the leave-one-out approach with denoising, are summarized in Table \ref{tab:comparison-w-denoise}. While the number of network parameters remains unchanged, accuracy improved across all models due to the noise reduction process. The Majority-Voting-based approach achieved a validation accuracy of 84.25\% and a test accuracy of 81.89\%, whereas our proposed network, incorporating wavelet denoising and leave-one-out validation, achieved a validation accuracy of 87.67\% and a test accuracy of 86.39\%. These results demonstrate that our model not only achieves higher accuracy but also maintains a lower computational cost, making it more suitable for real-time applications.

Tokas et al. \cite{Tokas2024} recently proposed a hybrid deep ensemble learning model that combines Convolutional Neural Networks (CNN) and Long Short-Term Memory (LSTM) architectures to recognize lower limb activities (sitting, standing, walking) from multichannel surface electromyography (sEMG) signals. Utilizing the same UCI sEMG dataset \cite{BASAN} as our work, their method achieved impressive classification accuracies—99.3\%, 98.3\%, and 98.8\% for walking, standing, and sitting activities respectively in healthy subjects, and 99.0\%, 98.1\%, and 98.2\% in pathological subjects. However, these results are derived from 3-fold cross-validation and not a leave-one-subject-out strategy, which is more appropriate for evaluating generalization in subject-independent scenarios. Additionally, while the authors do not report inference time, their model includes over 141,000 parameters, more than double the size of our model (62,876 parameters), implying a denser architecture with potentially higher computational demands that may hinder real-time deployment on edge devices. Furthermore, their approach relies on extrapolated signals via adaptive synthetic sampling (ADASYN) to balance class distribution. Although effective offline, such synthetic oversampling techniques are generally unsuitable for real-time systems, where incoming data cannot be artificially augmented. Thus, despite the strong offline performance, the model's applicability to real-time and embedded settings remains unverified.

%%%%%%%%%%%%%%%%%%%%%%%%%%%%%%%%%% TABLE
\begin{table*}[t!]
 \caption{Comparison of proposed network and baseline networks test and validation accuracy, together with the number of parameters for data without wavelet denoising.}
  \centering
  \begin{tabular}{lllll}
    \toprule
    % \multicolumn{4}{c}{Metrices}                   \\
    % \cmidrule(r){2-4}
    Network                              & Validation Accuracy (\%)       & Test Accuracy (\%)       & Numer of Parameters (\%) \\
    \midrule
    Conv1D-M1\cite{Vijayvargiya2021}       & 81.71                          & 78.15                    & 408,507   \\
    Conv1D-M2\cite{Vijayvargiya2021}       & 81.45                          & 78.67                    & 408,507   \\
    Conv1D-M3\cite{Vijayvargiya2021}       & 79.83                          & 78.02                    & 1,221,227 \\
    Conv1D-M4\cite{Vijayvargiya2021}       & 81.95                          & 81.61                    & 1,221,227 \\
    Majority-Voting\cite{Vijayvargiya2021} & 81.95                          & 81.61                    & Sumation of all four networks \\
    \textbf{Our Network}                   & \textbf{86.74}                 & \textbf{85.38}           & \textbf{62,876} \\
    \bottomrule
  \end{tabular}
  \label{tab:comparison-wout-denoise}
\end{table*}
%%%%%%%%%%%%%%%%%%%%%%%%%%%%%%%%%%%%%%%%%%%%%%%%%%%%%%%%%%%%%%%%%%%%%%%%%%%%
%%%%%%%%%%%%%%%%%%%%%%%%%%%%%%%%%% TABLE
\begin{table*}[t!]
 \caption{Comparison of proposed network and baseline networks test and validation accuracy, together with the number of parameters of each network for data with wavelet denoising.}
  \centering
  \begin{tabular}{lllll}
    \toprule
    % \multicolumn{4}{c}{Metrices}                   \\
    % \cmidrule(r){2-4}
    Network                              & Validation Accuracy (\%)       & Test Accuracy (\%)       & Numer of Parameters (\%) \\
    \midrule
    Conv1D-M1\cite{Vijayvargiya2021}       & 79.02                        & 81.89                  & 408,507   \\
    Conv1D-M2\cite{Vijayvargiya2021}       & 77.84                        & 78.12                  & 408,507   \\
    Conv1D-M3\cite{Vijayvargiya2021}       & 84.25                        & 80.31                  & 1,221,227 \\
    Conv1D-M4\cite{Vijayvargiya2021}       & 83.94                        & 78.87                  & 1,221,227 \\
    Majority-Voting\cite{Vijayvargiya2021} & 84.25                        & 81.89                  & Sumation of all four networks \\
    \textbf{Our Network}                   & \textbf{87.67}               & \textbf{86.39}         & \textbf{62,876} \\
    \bottomrule
  \end{tabular}
  \label{tab:comparison-w-denoise}
\end{table*}
%%%%%%%%%%%%%%%%%%%%%%%%%%%%%%%%%%%%%%%%%%%%%%%%%%%%%%%%%%%%%%%%%%%%%%%%%%%%

\section{Conclusion}

This research demonstrates that deep neural networks (DNNs) can be implemented as upper-level controllers in robotic systems. However, several challenges remain in fully realizing this goal. As observed in our results, uncertainties in class detection and misclassifications still exist, which, in a real-world scenario, could lead to undesired robotic actions. To mitigate these risks, it is essential to develop robust algorithms that can detect and address misclassifications effectively, ensuring greater reliability in robotic decision-making.

Beyond classification accuracy, an important factor remains largely unaddressed in prior research—including our own: the role of human perception and interaction with such upper-level controllers. A further investigation is necessary to understand how human users perceive and respond to these controllers, even in their current state with existing imperfections. It is also crucial to assess whether such an intelligent control mechanism enhances the overall performance and usability of the robotic system from the user’s perspective.

A possible starting point for this investigation could be testing the upper-level controller in a passive mode, allowing the robotic system to operate without directly influencing control decisions. This approach would help explore the interactive dynamics between humans and robots, providing deeper insights into how users adapt to and benefit from such AI-driven assistance in real-world scenarios.

In conclusion, this paper introduced an attention-based deep neural network with only 62,876 parameters, designed for the classification of lower limb sEMG signals. The proposed model achieved an accuracy of 86.74\% on the validation set (new subject) and 85.38\% on the test set under the leave-one-out condition, requiring only 50 epochs and no preprocessing. These results were obtained in a realistic scenario, where both the validation and test datasets consisted of previously unseen subjects, ensuring that the model was evaluated on truly novel data.

Furthermore, our model outperforms previous approaches in the literature, even when denoising filters are incorporated for data preprocessing. These findings highlight the efficiency and robustness of our approach, making it a promising candidate for real-time applications in robotic control, rehabilitation, and prosthetic systems.

\section{Future Works}
While this study demonstrates promising results for the classification of sEMG signals, several key aspects remain to be explored in future research. One important area of investigation is the impact of filtering on classification accuracy. As filtering has been shown to improve accuracy, a deeper analysis is needed to understand its effects on the data and to develop real-time implementable filtering techniques that enhance signal quality without introducing latency.

Another avenue for improvement lies in optimizing the network architecture through hyperparameter tuning methods, which could further refine the model’s performance and computational efficiency. Additionally, a critical challenge in real-world implementation of such upper-level controllers is managing transitions between movement classes, an aspect that has not been addressed in this study due to the lack of transition-phase data. Future research should focus on acquiring and analyzing transition-phase data to ensure smoother, more natural control in practical applications.

% \section{Acknowledgement}
% Thanks o Thanks :)
%%%%%%%%%%%%%%%%%%%%%%%%%%%%%%%%%%%%%%%%%%%%%%%%%%%%%%%%%%%%%%%%%%%%%%%%%%%%%%%%
\bibliographystyle{ieeetr}
% \IEEEtriggeratref{17} 
% \bibliography{references}

\end{document}